%% file: nnc-5.tex
\documentclass[11pt,twoside]{article}
\input epsf
\usepackage{natbib}
\usepackage{color}
\usepackage{epsfig}
\usepackage{pictex}

\newcommand{\psfigureabc}[6]{
        \begin{figure}[h!]
        \smallskip
        \begin{center}
	\begin{tabular}{ccc}
	        \epsfxsize=#5 \epsfbox{#2.eps} &
        	\epsfxsize=#5 \epsfbox{#3.eps} &
        	\epsfxsize=#5 \epsfbox{#4.eps} \\
		(a) & (b) & (c) \\
	\end{tabular}
        \end{center}
        \caption{#6}
        \label{fig:#1}
        \end{figure}
}

\begin{document}
\begin{center}
\Large
	On Interference of Signals and Generalization \\ in
	Feedforward Neural Networks
\vspace{0.2in}

\large
	Artur Rataj, e--mail arataj@iitis.gliwice.pl \\
	Institute of Theoretical and Applied Computer Science, \\
	Ba\l tycka 5, Gliwice, Poland

\vspace{0.2in}

\large
	Technical Report IITiS-2002-08-1-1.04
\normalsize
\vspace{0.2in}
\end{center}

\vspace{0.1in}

\begin{abstract}
This paper studies how the generalization ability of
neurons can be affected by mutual processing of different signals.
This study is done on the basis of a feedforward artificial
neural network.
The mutual processing of signals
can possibly be a good model of patterns in a set
generalized by a neural network
and in effect may improve generalization. In
this paper it is discussed that the interference may also cause
a highly random generalization. Adaptive activation functions
are discussed as a way of reducing that type of generalization. A test of
a feedforward neural network is performed that shows the discussed random
generalization.

\textbf{keywords:} feedforward
 neural networks, generalization, interference of signals, overfitting
\end{abstract}

\section{INTRODUCTION}
A feedforward artificial neural network, further denoted by FNN,
can be viewed as a rather `unconstrained' structure
-- in a typical multilayered architecture
an output of a neuron in one layer is simply connected to
all inputs in the succeeding layer, and the weights of connections
can just be initialized randomly. The combination function
of an artificial neuron of the \citet{mcculloch43nervous}
type treats all its arguments
as equivalent, simply adding them. In the process of training,
attributes of the training
observations are propagated through such a relatively generic structure,
possibly in a random order. It may rise several questions. How that
somewhat unconstrained structure of an artificial neural network copes
with generalization, especially when there are several
`competiting' stimuli, that simultaneously want to be `extrapolated' onto
`regions' in the inputs space of the FNN not covered by the training data.
How such conflicts can possibly destroy the ability of generalization, and what
can be the ways to reduce such phenomena?

\section{RANDOM GENERALIZATION}
The summing of signals in the combination function of an artificial neuron,
called here an \emph{interference} of signals, may improve generalization.
For example, in the case of a multi--dimensional data set, processing of values from
one input of a neural network can be influenced by values at another input of the
neural network, what may model well the patterns in the training set.
The error--minimizing
learning process can prevent harmful interference if the interference would
increase the neural network error of approximation of the training set. The
signals propagated from attributes of observations that are absent
in the training set, however, can be interfered with no effect on the error.
Therefore, the
interference can decrease the generalization ability of the network.
A decrease of generalization quality in neural networks can
also be an effect of overfitting \citep{schaffer91overfitting,
rosin95improving, lawrence97lessons, lawrence00overfitting}. Yet
the worsening of generalization caused by the discussed interference
can be very different  from that
caused by overfitting  While excessive fitting of the neural network
function to the training set means only that some particular patterns of
the set are memorized, the discussed interference of signals
may introduce \emph{highly random}
changes to the generalizing function of the neural network.

Let us further discuss such a type of a random generalization in more detail.

\section{STRONG PROPAGATION REGIONS}
\label{sec:propagation-regions}
In this section the so--called strong propagation regions in the input spaces of
neurons will be discussed. The notion will be used further in this paper to
describe the discussed interference of signals.

A neuron with linear weight functions and a hyperbolic tangent activation
function has its output value equal to a given
value \(r\) for its input values that, in the neuron
input space, create a hyperplane \(P_{r}\), except of
the special case where all weights in the neuron are equal to \(0\).
Specifically, there is a hyperplane
\(P_{0}\) for the neuron output value equal to \(0\). Because the hyperbolic
tangent activation functions have the greatest value of its derivative at \(0\),
the hyperplane \(P_{0}\) is the region in the neuron input space for which
there is the strongest propagation of signals through the neuron.  
As the distance from this hyperplane increases,
the derivative of the activation function decreases
and in effect the propagation becomes weaker. Let us call the region
with relatively strong level of propagation a strong propagation region.
Let the region consist of points whose distances to \(P_{0}\) in the
input space of the neuron do not exceed a certain value.

Let there be two fully connected subsequent layers \(L_{i}\) and
\(L_{i + 1}\) in a feedforward neural network. Let there be \(N_{i}\)
and \(N_{i + 1}\) neurons in the layers, respectively. Let us discuss the
input spaces of the neurons in the layer \(L_{i + 1}\). Each of the neurons
in the layer \(L_{i + 1}\) has \(N_{i} + 1\) inputs, \(N_{i}\) of which
are from the
neurons in the preceding layer and a single input is from the bias element.
Therefore, the transformation made in the layer \(L_{i + 1}\) can be
represented by parameterized \(N_{i + 1}\) \(N_{i}\)--dimensional input
spaces of the neurons in \(L_{i + 1}\), where the parameters in the spaces are
the values of functions of the respective neurons in \(L_{i + 1}\).

\begin{figure}[h!]
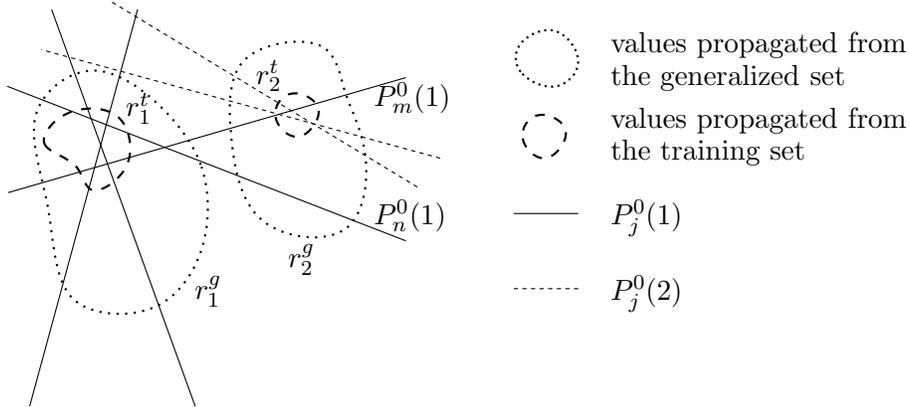

        \smallskip
        \begin{center}
	\input nn-interference-zero-places-3.pstex_t
        \end{center}
        \caption{An example diagram of input spaces of neurons in a layer.}
        \label{fig:nn-interference-zero-places-3}
        \end{figure}

An example of input spaces of neurons in \(L_{i + 1}\) is shown in
Fig.~\ref{fig:nn-interference-zero-places-3}.

The lines represent the hyperplanes \(P_{0}\), denoted by \(P^{0}_{j}\),
\(j = 0, 1, \ldots N_{i + 1} - 1\),
where \(j\) denotes a respective neuron in the layer \(L_{i + 1}\).
This is not a full representation of the input spaces of the neurons in the discussed
layer,
because the values of functions of the neurons
are not given, yet this diagram shows the regions
with the strong propagation of signals, being on and near the hyperplanes
\(P^{0}_{j}\).
The values propagated to the
neurons in the layer \(L_{i + 1}\) are either the direct values of attributes of
observations if \(L_{i + 1}\) is the first hidden layer, or images of the
attributes if \(L_{i + 1}\) is any of the succeeding layers. Anyway, the
region \(r_{t}\) of values propagated from the observations in the training set and
the region \(r_{g}\) of values propagated from
the observations in the generalized set can be shown in the input spaces of the neurons,
as it is done
in Figure \ref{fig:nn-interference-zero-places-3}.
In the example diagram, the region \(r_{t}\) consists of two regions
\(r^{t}_{p}\), \(p = 1, 2\), and the region \(r_{g}\) consists of another
two regions \(r^{g}_{q}\), \(q = 1, 2\). The regions are schematically
shown by solid regions in the diagrams, but they are
sets of discrete points, where each point corresponds to one or more observations.

Let each observation has its input attributes, that is these that are propagated
from the inputs of a neural network, and its output attributes, that is these that
are compared to values at the outputs of the network.
The hyperplanes \(P^{0}_{j}\)
in the example diagram generally concentrate in or near the regions \(r^{t}_{p}\).
This may happen during the training process
if there are relatively large differences between the values of output attributes
of observations whose input attributes are
propagated through \(r^{t}_{p}\). Thus,
relatively high values of derivatives of functions of the neurons in
\(L_{i + 1}\) may correspond to relatively large differences between the
output attributes of observations
in the training set. The hyperplanes \(P^{0}_{j}\), by
extending infinitely in the space, may allow for generalization to the points outside
\(r_{t}\), including the points that are relatively far from \(r_{t}\).

\section{INTERFERENCE OF SIGNALS}
\label{sec:interference-of-signals}
Let us discuss again the diagram of input spaces of neurons
in Figure \ref{fig:nn-interference-zero-places-3}.
Let there be several hyperplanes \(P^{0}_{j}\),
denoted by \(P^{0}_{j}(i)\), where \(j\) determines a respective neuron
and \(i = 1, 2\),
that were placed during the learning
process near \(r_{t}\), to minimize the component of \(\xi_{l}\) caused by
the observations in the training set, whose attributes propagate through \(r_{t}\).
They are marked in the diagram by solid lines for \(i = 1\) and by dotted lines
for \(i = 2\).
Let the regions \(r^{g}_{1}\) and \(r^{g}_{2}\) be overlapping or be near to
\(r^{t}_{1}\) or \(r^{t}_{2}\),
respectively. Let the observations whose input attributes are propagated through
the regions \(r^{g}_{1}\) and \(r^{g}_{2}\) be generalized well because of the
hyperplanes  \(P^{0}_{j}(1)\) and  \(P^{0}_{j}(2)\), respectively.
This is possible because the hyperplanes \(P^{0}_{j}(1)\) extend from \(r^{t}_{1}\)
and the hyperplanes
\(P^{0}_{j}(2)\) extend from \(r^{t}_{2}\), thus `extrapolating' the patterns
in the region \(r_{t}\). 

Now, if a hyperplane \(P^{0}_{j}(i)\), that normally is generalizing
patterns in \(r^{t}_{i}\), would by a chance `intersect' \(r^{t}_{3 - i}\),
like \(P^{m}_{0}(1)\) does,
it could possibly increase the training error \(\xi_{l}\),
and thus in a possible further training
the intersecting hyperplane \(P^{0}_{j}(i)\) could, for example,
be driven out of \(r^{t}_{3 - i}\). Yet if the
hyperplane would intersect \(r^{g}_{3 - i}\),
like \(P^{n}_{0}(1)\) does, it could intervene the
generalization from \(r^{t}_{3 - i}\) to \(r^{g}_{3 - i}\) without
any reaction in the training process. More, a region
\(r^{t}_{i}\) could, during the training, be placed itself in \(r^{g}_{3 - i}\),
thus causing all \(P^{0}_{j}(i)\), associated with generalization of
\(r^{t}_{i}\), to intervene the generalization to \(r^{g}_{3 - i}\).

The interference of signals,
causing a possibly high randomness of generalization,
could be reduced if the strong propagation region of a neuron
would not extend itself infinitely in space. This is
like in the radial basis function
neural networks \citep{broomhead1988, moody1989, poggio89theory}.
On the other hand, such forms of finite strong propagation
regions like in the radial basis function networks could worse the ability of
generalization of a neural network for sets where long
strong propagation regions are needed for good generalization.
A possible method of finding a good trade--off between infinite and finite
strong propagation regions could be using adaptive activation functions.
Such adaptive activation functions could, during training with a special
learning algorithm, smoothly adapt their form, for example in the range
between a radial basis function and a hyperbolic tangent. 

\section{TESTS}
\label{sec:tests}
Because in some relatively simple generalization
problems that were conducted
the discussed random generalization seemed to be rather rarely
observed -- usually the
trained neural networks after some time began only to overfit the data, showing
only some
randomness connected with a limited flexibility -- in this test a relatively
complex training set will be used.

\psfigureabc{training_set}
	{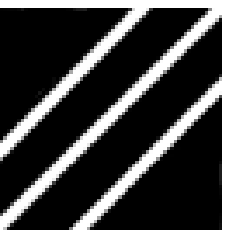}
	{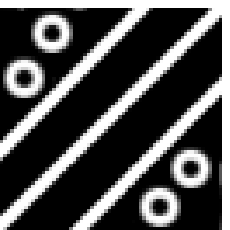}
	{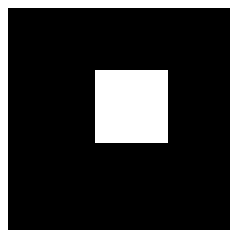}
	{0.7in}
	{The data sets (a) \(\theta_{l}\), (b) \(\theta_{c}\) and (c) the
	 training subsets mask.}
Let there be two three--dimensional sets \(\theta_{l}\) and
\(\theta_{c}\), as illustrated in
Figures \ref{fig:training_set}(a) and \ref{fig:training_set}(b),
respectively. The sets
are \(64\times 64\) images, whose pixel coordinates determine the neural network
input vector values, a single value for each dimension, and the pixels
brightnesses determine corresponding values in the neural network output vectors. 
The pixel at the lower left corner has the coordinates \((-0.5, -0.5)\) and the
pixel at the upper right corner has the coordinates \((0.5, 0.5)\). The brightness
of the pixels represents the range from \(-0.5\) for black to \(0.5\) for white.
Feedforward
layered networks with two inputs, a single neuron in the output layer and two hidden
layers of 16 neurons each, were trained by the training
subsets of either \(\theta_{l}\) or
\(\theta_{c}\). The neural networks had hyperbolic
tangent activation functions. There was a weight decay at a rate of
\(2\cdot 10^{-7}\) to improve generalization
\citep{krogh1992simple}. An online training was used with
a learning step of \(0.02\).
The training subsets are represented by the image in
Figure \ref{fig:training_set}(c). Black pixels in the image mean that the corresponding
pixels in Figures \ref{fig:training_set}(a) and \ref{fig:training_set}(b)
represent the training subsets of the respective generalized sets.

There were four neural networks \(\mathcal{N}^{l}_{i}\), \(i = 0 \ldots 3\),
trained with the subset of \(\theta_{l}\), and four another
neural networks \(\mathcal{N}^{c}_{i}\), \(i = 0 \ldots 3\), trained with the
subset of \(\theta_{c}\). The generalizing functions of the networks were
sampled and the weights of the neurons in the first input layer
were saved at each of the iterations 10000000th, 31622777th and 100000000th.
The results are illustrated in Fig.~\ref{fig:nnfigures}.
\newcommand{\nnfigures}[3] {
\begin{tabular}{cc}
	\epsfxsize=0.45in \epsfbox{mse.linedashed-#1-l7-ls0.#3.image.output.#2.eps} \\
	\epsfxsize=0.45in \epsfbox{mse.linedashed-#1-l7-ls0.#3.image.l1s.#2.eps} \\
\end{tabular}
\hspace{-0.35in}
}
\begin{figure}[h!]
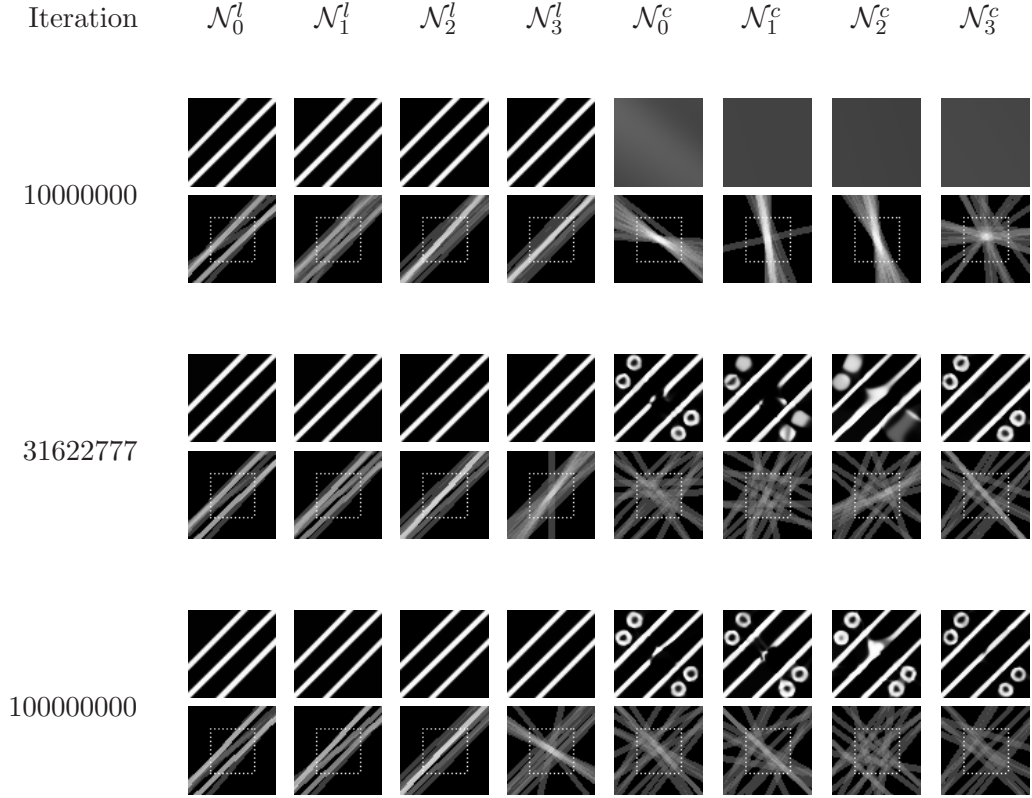

\smallskip
\begin{center}
\begin{tabular}{rcccccccc}
		Iteration &
		\(\hspace{0.20in} \mathcal{N}^{l}_{0}\) & \(\hspace{0.20in} \mathcal{N}^{l}_{1}\) &
		\(\hspace{0.20in} \mathcal{N}^{l}_{2}\) & \(\hspace{0.20in} \mathcal{N}^{l}_{3}\) &
		\(\hspace{0.20in} \mathcal{N}^{c}_{0}\) & \(\hspace{0.20in} \mathcal{N}^{c}_{1}\) &
		\(\hspace{0.20in} \mathcal{N}^{c}_{2}\) & \(\hspace{0.20in} \mathcal{N}^{c}_{3}\) \hspace{0.35in} \\
		\\[10pt]
		10000000 &
		\nnfigures{7}{10000000}{0} &
		\nnfigures{7}{10000000}{1} &
		\nnfigures{7}{10000000}{2} &
		\nnfigures{7}{10000000}{3} &
		\nnfigures{6}{10000000}{0} &
		\nnfigures{6}{10000000}{1} &
		\nnfigures{6}{10000000}{2} &
		\nnfigures{6}{10000000}{3} \hspace{0.35in} \\
		\\[10pt]
		31622777 &
		\nnfigures{7}{31622777}{0} &
		\nnfigures{7}{31622777}{1} &
		\nnfigures{7}{31622777}{2} &
		\nnfigures{7}{31622777}{3} &
		\nnfigures{6}{31622777}{0} &
		\nnfigures{6}{31622777}{1} &
		\nnfigures{6}{31622777}{2} &
		\nnfigures{6}{31622777}{3} \hspace{0.35in} \\
		\\[10pt]
		100000000 &
		\nnfigures{7}{100000000}{0} &
		\nnfigures{7}{100000000}{1} &
		\nnfigures{7}{100000000}{2} &
		\nnfigures{7}{100000000}{3} &
		\nnfigures{6}{100000000}{0} &
		\nnfigures{6}{100000000}{1} &
		\nnfigures{6}{100000000}{2} &
		\nnfigures{6}{100000000}{3} \hspace{0.35in} \\
\end{tabular}
\end{center}
\caption{The generalizing functions and diagrams 
of the zeroes of the first hidden layer neurons.}
\label{fig:nnfigures}
\end{figure}
There is a table for each iteration in the figure, with sampled generalization functions
in the upper row and diagrams representing input spaces of neurons
in the first hidden layer in the lower row.
The representation of the generalization functions
is analogous to that of the sets \(\theta_{l}\) and \(\theta_{c}\).
Each of the input space diagrams shows with
translucent lines the zeroes of the outputs of the first hidden layer neurons,
that is it shows the hyperplanes \(P^{0}_{j}\),
against the common input values from the input layer.
The lower left corner of the dotted
rectangles drawn within the diagrams represents input values \((-0.5, -0.5)\)
and the upper right corner of the rectangles represents input values
\((0.5, 0.5)\). Therefore, the input attributes of the observations in the sets
\(\theta_{l}\) and \(\theta_{c}\) are propagated into the space marked
in the diagrams by the dotted rectangles. The propagation to the first hidden layer
is without any
transformation of course, because the nodes in the input layer only
pass signals to the first hidden layer.

Let us look at the diagrams of the input spaces of the neurons in the first hidden layer.
Because of the direct relation between the space of the
input attributes of the observations and
the input spaces of the first hidden layer neurons it can be said that
in the cases of both \(\mathcal{N}^{l}_{i}\) and  \(\mathcal{N}^{c}_{i}\)
the hyperplanes \(P^{0}_{j}\) generally concentrate as it was discussed
in Sec.~\ref{sec:propagation-regions}.
In particular, in
\(\mathcal{N}^{c}_{i}\), generally some hyperplanes concentrate near
the linear features \(f_{l}\) and some concentrate near the circular features
\(f_{c}\). In effect, the lines in the diagrams
concentrated near \(f_{c}\) cross these concentrated near \(f_{l}\).
Additionally, the crossings occur partially
in the region not covered by the training set. These are exactly the conditions
prone to the random generalization, discussed in
Sec.~\ref{sec:interference-of-signals}. In fact, unlike
\(\mathcal{N}^{l}_{i}\), where the hyperplanes finely `extrapolate' the regions
in the training file, in the functions of \(\mathcal{N}^{c}_{i}\)
a highly random generalization can be seen.

\section{CONCLUSIONS}
\label{sec:conclusions}
It was discussed that the interference of signals within a
FNN, while possibly being one of its strengths, may
also cause a substantially random generalization.
Tests of generalization of two sets of data was presented.
The obtained generalizing function was relatively predictable
in the case of one of the sets, and there was a high randomness
in the function in the case of the other set.

\bibliography{nn}
\bibliographystyle{apalike}

\end{document}

%% file: nn-interference-zero-places-3.pstex_t
\begin{picture}(0,0)%
\includegraphics{nn-interference-zero-places-3.pstex}%
\end{picture}%
\setlength{\unitlength}{3552sp}%
\begingroup\makeatletter\ifx\SetFigFont\undefined%
\gdef\SetFigFont#1#2#3#4#5{%
  \reset@font\fontsize{#1}{#2pt}%
  \fontfamily{#3}\fontseries{#4}\fontshape{#5}%
  \selectfont}%
\fi\endgroup%
\begin{picture}(6056,2861)(2753,-5619)
\put(6976,-3136){\makebox(0,0)[lb]{\smash{\SetFigFont{11}{13.2}{\rmdefault}{\mddefault}{\updefault}{\color[rgb]{0,0,0}values propagated from}%
}}}
\put(6976,-3661){\makebox(0,0)[lb]{\smash{\SetFigFont{11}{13.2}{\rmdefault}{\mddefault}{\updefault}{\color[rgb]{0,0,0}values propagated from}%
}}}
\put(6976,-3886){\makebox(0,0)[lb]{\smash{\SetFigFont{11}{13.2}{\rmdefault}{\mddefault}{\updefault}{\color[rgb]{0,0,0}the training set}%
}}}
\put(3594,-3579){\makebox(0,0)[lb]{\smash{\SetFigFont{11}{13.2}{\rmdefault}{\mddefault}{\updefault}{\color[rgb]{0,0,0}\(r^{t}_{1}\)}%
}}}
\put(4501,-3354){\makebox(0,0)[lb]{\smash{\SetFigFont{11}{13.2}{\rmdefault}{\mddefault}{\updefault}{\color[rgb]{0,0,0}\(r^{t}_{2}\)}%
}}}
\put(4074,-4854){\makebox(0,0)[lb]{\smash{\SetFigFont{11}{13.2}{\rmdefault}{\mddefault}{\updefault}{\color[rgb]{0,0,0}\(r^{g}_{1}\)}%
}}}
\put(6976,-3361){\makebox(0,0)[lb]{\smash{\SetFigFont{11}{13.2}{\rmdefault}{\mddefault}{\updefault}{\color[rgb]{0,0,0}the generalized set}%
}}}
\put(5326,-3511){\makebox(0,0)[lb]{\smash{\SetFigFont{11}{13.2}{\rmdefault}{\mddefault}{\updefault}{\color[rgb]{0,0,0}\(P^{0}_{m}(1)\)}%
}}}
\put(5326,-4336){\makebox(0,0)[lb]{\smash{\SetFigFont{11}{13.2}{\rmdefault}{\mddefault}{\updefault}{\color[rgb]{0,0,0}\(P^{0}_{n}(1)\)}%
}}}
\put(4726,-4636){\makebox(0,0)[lb]{\smash{\SetFigFont{11}{13.2}{\rmdefault}{\mddefault}{\updefault}{\color[rgb]{0,0,0}\(r^{g}_{2}\)}%
}}}
\put(6976,-4861){\makebox(0,0)[lb]{\smash{\SetFigFont{11}{13.2}{\rmdefault}{\mddefault}{\updefault}{\color[rgb]{0,0,0}\(P^{0}_{j}(2)\)}%
}}}
\put(6976,-4336){\makebox(0,0)[lb]{\smash{\SetFigFont{11}{13.2}{\rmdefault}{\mddefault}{\updefault}{\color[rgb]{0,0,0}\(P^{0}_{j}(1)\)}%
}}}
\end{picture}

%% file: nnc-5.bbl
\begin{thebibliography}{}

\bibitem[Broomhead and Lowe, 1988]{broomhead1988}
Broomhead, D.~S. and Lowe, D. (1988).
\newblock Multivariable functional interpolation and adaptive networks.
\newblock {\em Complex Systems}, 2:321--355.

\bibitem[Krogh and Hertz, 1992]{krogh1992simple}
Krogh, A. and Hertz, J.~A. (1992).
\newblock A simple weight decay can improve generalization.
\newblock In Moody, J.~E., Hanson, S.~J., and Lippmann, R.~P., editors, {\em
  Advances in Neural Information Processing Systems}, volume~4, pages 950--957.
  Morgan Kaufmann Publishers, Inc.

\bibitem[Lawrence and Giles, 2000]{lawrence00overfitting}
Lawrence, S. and Giles, C.~L. (2000).
\newblock Overfitting and neural networks: Conjugate gradient and
  backpropagation.
\newblock In {\em Proceedings of the {IEEE} International Conference on Neural
  Networks}, pages 114--119. IEEE Press.

\bibitem[Lawrence et~al., 1997]{lawrence97lessons}
Lawrence, S., Giles, C.~L., and Tsoi, A.~C. (1997).
\newblock Lessons in neural network training: Overfitting may be harder than
  expected.
\newblock In {\em Proceedings of the Fourteenth National Conference on
  Artificial Intelligence, {AAAI}-97}, pages 540--545. AAAI Press, Menlo Park,
  California.

\bibitem[McCulloch and Pitts, 1943]{mcculloch43nervous}
McCulloch, W.~S. and Pitts, W.~H. (1943).
\newblock A logical calculus of the ideas immanent in nervous activity.
\newblock {\em Bulletin of Mathematical Biophysics}, 5:115--133.

\bibitem[Moody and Darken, 1989]{moody1989}
Moody, J. and Darken, C. (1989).
\newblock Fast learning in networks of locally tuned units.
\newblock {\em Neural Computations}, 1(2):281--294.

\bibitem[Poggio and Girosi, 1989]{poggio89theory}
Poggio, T. and Girosi, F. (1989).
\newblock A theory of networks for approximation and learning.
\newblock Technical Report AIM-1140.

\bibitem[Rosin and Fierens, 1995]{rosin95improving}
Rosin, P. and Fierens, F. (1995).
\newblock Improving neural network generalisation.

\bibitem[Schaffer, 1991]{schaffer91overfitting}
Schaffer, C. (1991).
\newblock Overfitting avoidance as bias.
\newblock In {\em {IJCAI}-91 Workshop on Evaluating and Changing Representation
  in Machine Learning}, Sydney.

\end{thebibliography}
